\documentclass[conference]{IEEEtran}
\IEEEoverridecommandlockouts
\usepackage{cite}
\usepackage{amsmath,amssymb,amsfonts}
\usepackage{algorithmic}
\usepackage{textcomp}
\usepackage{xcolor}
\usepackage{booktabs}
\usepackage{CJKutf8} % use CJK to typeset Chinese, Japanese and Korean
\usepackage{multirow}
\usepackage{makecell}
\usepackage{tikz}
\usepackage{pgfplots}
\usepackage{xcolor} % 加载 xcolor 宏包，用于控制颜色
\def\BibTeX{{\rm B\kern-.05em{\sc i\kern-.025em b}\kern-.08em
    T\kern-.1667em\lower.7ex\hbox{E}\kern-.125emX}}
\DeclareRobustCommand*{\IEEEauthorrefmark}[1]{%
    \raisebox{0pt}[0pt][0pt]{\textsuperscript{\footnotesize\ensuremath{#1}}}}
\begin{document}

\title{TCM-GPT: Efficient Pre-training of Large Language Models for Domain Adaptation in Traditional Chinese Medicine\\
% 标题候选
% TCM-Adapt: Fine-Tuning Large Language Models for Domains of Traditional Chinese Medicine
% TCM-GPT: Tailoring Large Pretrained Language Models for Specialized Domains in Traditional Chinese Medicine
% TCM-GPT:  Efficient Pre-training of Large  Language Models for Specialized Domains of Traditional Chinese Medicine
% TCMDA: Large Language Models Domain Adaptation in Traditional Chinese Medicine
% TCMDALLM: A Pre-trained Large Language Model for TCM by Domain Adaptation
% Efficient Pre-training of Large Language Models for Domain Adaptation in Traditional Chinese Medicine

% {\footnotesize \textsuperscript{*}Note: Sub-titles are not captured in Xplore and should not be used}
}

\author{
\IEEEauthorblockN{
Guoxing Yang\IEEEauthorrefmark{1},
Jianyu Shi\IEEEauthorrefmark{1},
Zan Wang\IEEEauthorrefmark{1},
Xiaohong Liu\IEEEauthorrefmark{2}$^{,*}$,
Guangyu Wang\IEEEauthorrefmark{1}$^{,*}$}
\IEEEauthorblockA{\IEEEauthorrefmark{1}
\textit{State Key Laboratory of Networking and Switching Technology}, \\ 
\textit{Beijing University of Posts and Telecommunications, Beijing 100876, China}}
\IEEEauthorblockA{\IEEEauthorrefmark{2}
\textit{UCL Cancer Institute, University College London, London, United Kingdom}}

\IEEEauthorblockA{
yangguoxing@bupt.edu.cn, shi\_jian\_y@163.com, wangz@bupt.edu.cn, xhliu17@gmail.com, guangyu.wang@bupt.edu.cn
}
\IEEEauthorblockA{*Corresponding Author: Xiaohong Liu, Guangyu Wang \quad Email: xhliu17@gmail.com, guangyu.wang@bupt.edu.cn}
}
\maketitle

% In this paper, we constructed a TCM corpus by extracting keywords from task data and retrieving a TCM subset corpus from general corpus.
% I believe that the corpus we constructed can be helpful for subsequent studies in the field of TCM.
% We conducted pre-training on this TCM corpus to achieve domain adapation in the filed of TCM.
% To save computational resources, we employed the LoRa to train.
% We conduct a multi-level evaluation of our approach, including   simple TCM examination task and complex real-world TCM EHR diagnosis task.
% Overall, our method contributes to improving the performance of models on various downstream tasks in the field of Traditional Chinese Medicine. 
% 创新点 域适应
% 首次在中医领域上验证，验证了领域适配
% 1. 我们构建一套方法，提取领域语料，可以提高预训练效率。中医预训练语料的构建
% 2. 为了提取"领域相关"的语料，我们提出 领域 关键词加权 检索。预训练的时候，引入LoRA，减少领域适配的参数量，保留通用能力，减少过拟合风险
% 3. 结果显示，中医领域两个任务，都很好
% 4. 说之前是小模型，现在比较大，之前是bert-encoder，现在是生成结构，decoder.
% 5. 我们接受了会放出模型

\begin{abstract}
% 预训练微调已经成为NLP的范式
Pre-training and fine-tuning have emerged as a promising paradigm across various natural language processing (NLP) tasks. The effectiveness of pretrained large language models (LLM) has witnessed further enhancement, holding potential for applications in the field of medicine, particularly in the context of Traditional Chinese Medicine (TCM).
% 在一些特定领域，例如医学，尤其是更为特定的领域如中医学，预训练语料中缺乏针对性的训练语料，这可能导致知识缺乏，增加了产生虚假信息或幻觉的风险。
% In certain domains, such as medicine, especially Traditional Chinese Medicine (TCM), there is a limited availability of domain-specific pre-training corpus. This can lead to a lack of knowledge and an increased risk of generating false information or illusions.
However, the application of these general models to specific domains often yields suboptimal results, primarily due to challenges like lack of domain knowledge, unique objectives, and computational efficiency. Furthermore, their effectiveness in specialized domains, such as Traditional Chinese Medicine,  requires comprehensive evaluation.
% 因此，用一些特定领域的数据去做领域适配是非常有必要的。
% Therefore, it is highly necessary to perform domain adaptation using specific domain data.

% 但是中医领域往往缺乏大规模的领域语料库，很难去做领域适配
% However, it is difficult to find a large amount of domain-specific corpus for TCM.
% 我们构建一套方法，提取领域语料，可以提高预训练效率。中医预训练语料的构建
% 训练的时候，引入LoRA，减少领域适配的参数量
% To overcome the limited availability of pre-training data in TCM, we propose a strategic method called TCM Domain Adapatation (TCMDA) to construct a corpus in TCM called TCM-Corpus-1B and adapt language model on it. 
% We utilize the LoRA technique for domain adaptation during pre-training and fine-tuning to align large language models to TCM domain, and the trained model is named TCM-GPT-7B.
% We conduct a multi-level evaluation of TCM-GPT-7B on TCM examination and TCM diagnosis.
% The experimental results demonstrate that our models achieve significantly higher performance in various TCM tasks after pre-training with the TCM-Corpus-1B, indicating that TCMDA is necessary and effective.
To address the above issues, we propose a novel domain specific TCMDA (TCM Domain Adaptation) approach, efficient pre-training with domain-specific corpus. Specifically, we first construct a large TCM-specific corpus, TCM-Corpus-1B, by identifying domain keywords and retreving from general corpus. Then, our TCMDA leverages the LoRA which freezes the pretrained model's weights and uses rank decomposition matrices to efficiently train specific dense layers for pre-training and fine-tuning, efficiently aligning the model with TCM-related tasks, namely TCM-GPT-7B. We further conducted extensive experiments on two TCM tasks, including TCM examination and TCM diagnosis. TCM-GPT-7B archived the best performance across both datasets, outperforming other models by relative increments of 17\% and 12\% in accuracy, respectively. 
To the best of our knowledge, our study represents the pioneering validation of domain adaptation of a large language model with 7 billion parameters in TCM domain.
We will release both TCM-Corpus-1B and TCM-GPT-7B model once accepted to facilitate interdisciplinary development in TCM and NLP, serving as the foundation for further study.

\end{abstract}


\section{Introduction}
% 预训练和大模型
% 预训练模型，取得很好的效果
% 进一步的，近年来，更大的预训练语言模型，进一步取得很好效果
In recent years, the paradigm of pre-training language models followed by fine-tuning has gained significant attention in the field of Natural Language Processing (NLP). Based on a wide variety of textual content, pretrained language models (PLMs) have shown their capacit in capturing deep semantic insights, significantly enhancing performance across numerous NLP tasks \cite{han2021pre}. 
For instance, BERT \cite{alaparthi2020bidirectional} employs stacked Transformer encoders for comprehensive corpus pre-training, resulting in marked performance gains in downstream tasks. Concurrently, GPT \cite{radford2018improving} adopts a Transformer-based decoder architecture, illustrating notable generative capabilities.
Recent studies indicate that by scaling up both the model size and training data, large language models (LLMs) can achieve impressive results \cite{zhao2023survey}. For example, models such as GPT-3 \cite{brown2020language} amassing 175 billion parameters, and PaLM \cite{chowdhery2022palm} with its remarkable 540 billion parameters, have been trained on datasets with over a trillion tokens. These datasets include a diverse range of sources, from programming code snippets to news articles, forum threads, and books. Through comprehensive training, these models demonstrate emergent behaviors, including few-shot learning and zero-shot learning, enabling them to excel in tasks such as writing, programming, and reasoning.

% 大模型基座上的预训练
% 预训练大模型在两方面的探索：
% 1 算力，效率方面的问题，一些先前的方法
% 2 不同领域或任务适配；
Despite the impressive performance of large language models, their direct application to specific domains encounters challenges, including complex domain knowledge, unique objectives, and  computational efficiency.
% , which hinders many budget-constrained research groups from pre-training their own language models and conducting explorations and improvements
A primary hurdle is the challenge faced by a general LLM in incorporating the specialized domain-specific knowledge without pretraining, which can cause performance degradation when adapting models to diverse contexts. The lack of domain knowledge can even lead to the generation of hallucination or false information. Gururangan et al. \cite{gururangan2020don} demonstrated that domain-adaptive pretrained models consistently outperformered that without pretraining, underscoring the indispensability of pretraining, when evaluated on four domains, including biomedical, computer science, news, and reviews. Moreover, Arumae et al. \cite{arumae2020empirical} and Jin et al. \cite{jin2021lifelong} found that parameter regularization techniques for pretrained models can yield superior results in domain adaptation. 
Shah, R. S., et al.\cite{shah2022flue} presents FLANG, a domain-specific language model tailored for financial contexts. By utilizing financial keywords and phrases, it optimizes its training process and offers benchmarks for five unique NLP tasks within the financial domain.
Another challenge is the computational cost of large-scale pre-training, which is often prohibitively high due to large-scale parameter size or stacked network layers. To address this, ALBERT \cite{lan2019albert} employs cross-layer parameter sharing in the Transformer architecture, which substantially reduces the model size. However, this leads to an obvious performance degradation. DistilBERT \cite{sanh2019distilbert} serves as a distilled version of large pretrained models, reducing parameters with maintaining commendable performance. Edward et al. \cite{hu2021lora} introduced the Low-Rank Adaptive (LoRA) approach, wherein they freeze the pretrained model's weights and use rank decomposition matrices to efficiently train specific dense layers, resulting in significantly reducing the number of trainable parameters without compromising performance on subsequent tasks.

% 预训练大模型在一些生物医学领域上的应用。
% 在中医上的挑战
In the field of biomedicine, there exist research that leverages pre-trainedlanguage models in specialized domains, demonstrating promising results.
% BioBERT  生物医学领域预训练，微调以应用于多个下游任务。但是模型比较小
% BioMedLM 
For instance, BioBERT \cite{lee2020biobert}, a domain-specific BERT-based model pre-trained on biomedical text, has demonstrated applicability across downstream tasks including biomedical named entity recognition and relation extraction. Venigalla et al. \cite{BioMedLM2023} introduced BioMedLM, a GPT-based mode trained on the PubMed abstracts and full text documents, achieving state-of-the art results on medical question and answer tasks. Despite the existing research of pre-trained language models, few studies have explored scaling them within the clinical domain, due to the sensitive nature of clinical narratives and the substantial computing power required. Furthermore, their effectiveness in specialized domains, such as Traditional Chinese Medicine (TCM), remains to be validated. 

In this study, we aim to investigate the adaptation of large language models to TCM domain and subsequently conduct performance evaluations. This effort carries considerable significance, as Traditional Chinese Medicine represents a cherished heritage that has been passed down through generations in China, presenting a noteworthy complement to Western medicine \cite{tang2008traditional}.
% 中医领域之前已有的一些相关工作
% A Question-Answering System over Traditional Chinese Medicine - 2015
% Traditional Chinese Medicine Language System (TCMLS)
There have been a few studies focused on NLP tasks in TCM.
For example, Zhang et al. \cite{zhang2020constructing} proposed the Traditional Chinese Medicine Language System (TCMLS), aiming to standardize TCM terminology to enhance information extraction. Nevertheless, constructing such a system incurred high data acquisition costs.
% 尽管有这些相关的应用，在与训练大模型上中医仍是缺失的，存在一些挑战，具体
% 1 已有预训练模型中，通常没有用到中医语料，可能导致或恶化幻觉问题
% 2 中医公开资源更少
% 3 领域模型的计算资源通常比较少
However, the integration of large language models into the TCM is currently not well developed compared to Western medicine. This discrepancy can be attributed to a range of challenges. Firstly, in contrast to other medical domains, Traditional Chinese Medicine faces a scarcity of publicly available resources. Many pre-trained models are predominantly fine-tuned on generalized corpora, potentially leading to suboptimal performance when confronted with the nuanced language of TCM
\cite{yin2023large,he2022rethinking}. Moreover, the computational resources available for training domain-specific models are generally scarcer compared to the abundance available for generic models.

% 本文的方法 (需要加上一些关键点：大模型、第一次中医)
% 首次在验证大模型在中医的领域适配。(说之前是小模型，现在比较大，之前是bert-encoder，现在是生成结构，decoder.)
% 1. 我们构建一套方法，提取领域语料，可以提高预训练效率。中医预训练语料的构建
% 2. 为了提取"领域相关"的语料，我们提出 领域 关键词加权 检索。预训练的时候，引入LoRA，减少领域适配的参数量，保留通用能力，减少过拟合风险。结果显示，中医领域对两个任务，无论侧重医学知识还是临床实践，都很好 
% 3. 最后，我们会放出模型TCMDA和中医语料
To address these challenges, we explored the process of  tailoring general-purpose LLMs to the specific TCM domain through the integration of contextual data and domain-specific knowledge. We introduced a novel approach, TCMDA, aimed at efficiently customizing pre-training models for the TCM field.
Our contributions are summarized as follows:
Firstly, to overcome the limited availability of pre-training data in TCM, we adopted a strategic approach by extracting domain text from general corpus. By incorporating domain terminologies and extracting task-relevant keywords, we constructed a set of keywords specific to TCM. This set was then utilized to retrieve TCM-relevant text, resulting in the TCM-Corpus-1B, a substantial collection of TCM-related textual data.
Secondly, to ensure the alignment of large language models to the TCM domain, we introduced a novel model named TCM-GPT-7B. This model adopts the LoRA technique for domain adaptation during both pre-training and fine-tuning phases. Furthermore, we performed a comprehensive evaluation of TCM-GPT-7B across two distinct TCM tasks: TCM examinations and TCM diagnosis. The experimental results demonstrate our models archive state-of-the-art performance after pre-training using the TCM-Corpus-1B, revealing the necessity and effectiveness of domain-specific pre-training. To our knowledge, TCM-GPT-7B represents a pioneer as the first 7B-scale large language model pre-trained exclusively on TCM data.
Finally, we are committed to publicly releasing both the curated the TCM-Corpus-1B and the TCM-GPT-7B model, to advance the cross-disciplinary research in both TCM and NLP.

\begin{figure*}[th]
  \centering
  \includegraphics[width=1.0\textwidth]{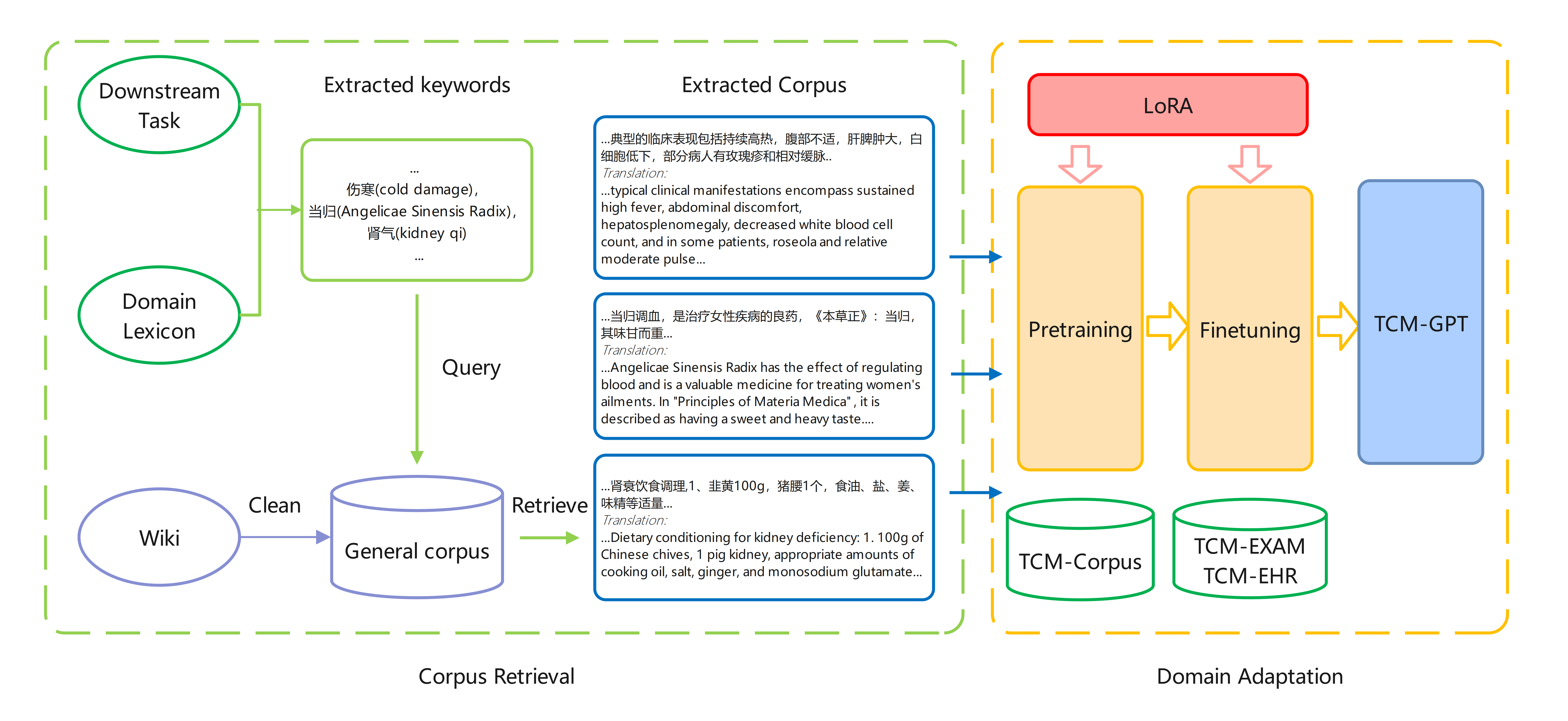} % 替换为你的图片文件名和路径
  \caption{The overview of TCMDA}
  \label{fig:1}
\end{figure*}

\section{Methods}

% 总体上说一下方法。
In this section, we describe our proposed TCMDA. Our method can be divided into three steps: domain-related keyword extraction, associated corpus retrieval, and LLM pretraining and fine-tuning, as shown in Figure \ref{fig:1}.

\subsection{Domain-related Keyword Extraction}
% 我们使用TextRank算法来提取关键词。
We utilized the TextRank algorithm \cite{mihalcea2004textrank} to extract a set of keywords related to task, denoted as $K^T=\{k_1,\cdots,k_n\}$, where $k_i \in K^T$ is a keyword, $n$ represents the number of keywords. 
TextRank is a graph-based ranking algorithm, used for automates keyword extraction in text. It builds a graph by measuring word similarity, employing iteration to assess word importance and derive keyword rankings.
By applying TextRank to the text extracted from all samples, we acquired the top five keywords for each individual sample.
To augment keyword diversity and enhance the domain coverage, we incorporated a publicly accessible TCM lexicon\footnote{https://pinyin.sogou.com/dict/cate/index/135} into the keyword collection, denoted as $K^L$. Consequently, this fusion of both $K^T$ and $K^L$ led to the formation of the comprehensive keyword set, $K^D$, which served as the foundation for querying the corpus.

Furthermore, during the keyword extraction process for each sample, we incorporated quantity information to enhance the effectiveness of retrieving task-relevant corpus. Specifically, we employed the count of keyword occurrences as weights when retrieval. The weight was computed as follows:
\begin{align}
    w_i ={1 + {\log(n_i)}},
\end{align}
% \begin{align}
%     w_i = \frac{1 + \lfloor{\log(n_i)}\rfloor}{\sum_{i=1}^{n}(1 + \lfloor{\log(n_i)}\rfloor)},
% \end{align}
where $n_i$ represents the count of keyword $k_i$.
%$n$ is the number of unique keywords.

\subsection{Associated Corpus Retrieval}
% 交代一下用的检索算法BM25。
% Although there are other retrieval methods with better performance, t
% The BM25 algorithm is sufficient to verify the effectiveness of our method.
% BM25 是一种用来评价搜索词和文档之间相关性的算法，它是一种基于概率检索模型提出的算法。

% corpus来源于百度百科和维基百科
We collected an extensive language corpus from both Baidu Baike\footnote{https://baike.baidu.com/} and  Wikipedia sources\footnote{https://zh.wikipedia.org/}. Before extracting retrieval from the corpus, a thorough data cleaning process was performed. 
We have removed the special tags and characters, and standardized the  format for retrieved contents.
% 交代一下用的检索算法BM25。
To effectively retrieve domain-specific content from the collected corpus, we employed the BM25 algorithm \cite{robertson2009probabilistic}, as it's simple, efficient, and widely used in information retrieval tasks\cite{yao2022nlp}. BM25 is based on the TF-IDF (Term Frequency-Inverse Document Frequency) model. It treats text as a collection of terms and calculates a score $score(D,Q)$ between a given document $D$ from database and a query $Q$. 
To enhance the efficacy of BM25, we adjusted a query keyword's weight by a previous obtained weight $w_i$. Specifically, we repeated a keyword $k_i$ with $\min(\lfloor w_i \rfloor, 3)$ times. The resultant modified query $Q'$ is then used to score the documents, ultimately yielding the final relevance score $score(D,Q')$.

\subsection{LLM Pretraining and Fine-tuning with LoRA}
To address the challenges associated with excessive parameter size and the risk of overfitting, we adopted the LoRA approach, we employed the LoRA \cite{hu2021lora} to for both pre-training and fine-tuning, as an alternative to conventional full-parameter fine-tuning strategies.
The LoRA methodology involves the freezing of pretrained model matrix weights $W$, 
coupled with the introduction of trainable rank decomposition matrices $B$ and $A$ into each layer of the Transformer architecture, greatly reducing the number of trainable parameters for downstream tasks. Specifically, given a linear layer $h=Wx$, the LoRA transforms this layer into $h=(W + W^{\prime})x$, where $W$ represents frozen parameters, $W^{\prime}=BA$ represents trainable parameters, $B \in \mathbb{R}^{d_1 \times r}$, $A \in \mathbb{R}^{r \times d_2}$, and the rank $r \ll \min(d_1,d_2)$.
During pre-training process, we optimize a causal language model (CLM) loss function designed to predict the subsequent token:
\begin{align}
L_{CLM} = -\frac{1}{n}\sum_{i=1}^{n} \log P_{{\theta}^{\prime}}(w_i|w_{1}, ..., w_{i-1})
\end{align}
where $w_{1},\cdots,w_{i-1}$ denotes the preceding $i-1$ tokens in a sentence, $w_i$ denotes the next token.
During instruction supervised fine-tuning (SFT) process, for a given input prompt $I=w_{1:m}$ and response $R=w_{m+1:m+n}$, the loss function is:
\begin{align}
L_{SFT} & = -\log P_{\theta^{\prime\prime}} (R|I) \\
           & = -\frac{1}{n}\sum_{i=m+1}^{m+n}{\log P_{\theta^{\prime\prime}} (w_i|w_1,...,w_{m-1},w_{m},..,w_{i-1})}
\end{align}
where $n$ and $m$ represent the lengths of the response and input prompt, respectively.

\section{Experiments}
\subsection{Datasets}
% 数据包括3个：encyclopedic corpus, TCM examination data, and TCM diagnostic data
The data utilized in this study primarily consists of three components: a general corpus, a TCM examination dataset (denoted as TCM-EXAM), and a TCM Electronic Health Record dataset (denoted as TCM-EHR).
The general corpus was collected from Baidu Baike and Wikipedia sources. After undergoing a  data cleaning process, this corpus encompasses a substantial volume of information, totally 20GB in size.
% TCM examination data 考试数据是由中国不同来源的考卷构成, 它总共包含了6000多道单选题。
The TCM-EXAM dataset comprises TCM-specific examination questions from diverse sources. It encompasses a total of 6,325 multiple-choice questions. Within this dataset, we constructed a test subset that concentrates on questions related to Diagnostics (denoted as DT) and Formula Science (denoted as FS) in TCM, containing 214 and 207 questions, respectively. The remaining 5,904 questions are used as the train set. 
% TCM diagnostic data 是我们从一家医院的信息科收集来的 
The TCM-EHR dataset is collected from a traditional Chinese medicine hospital and encompasses 7,783 electronic health records. Each individual record encapsulates information including chief complaints, history of present illness, physical examinations, and diagnoses, among others. For this dataset, a targeted test subset was generated, including specific disease groups, namely "Etiology and syndrome" (denoted as ES) and "Skin and mucosal diseases" (denoted as SMD) within TCM. This test subset encompasses 150 and 150 records, while the remaining 7,483 records are used as the training set.

\subsection{Evaluation Metric}
% 我们在两个任务上评价模型的性能：中医考试任务和中医诊断任务
We conducted performance evaluations for our model across two distinct tasks: the TCM examination task using the TCM-EXAM dataset, and the TCM diagnosis task using the TCM-EHR dataset.
% 对于中医考试任务，我们直接用考试题目。
For the TCM examination task, we employed the examination questions along with their corresponding answer choices as input. We then extracted the predicted option from the model-generated response using regular expressions.
% 对于中医诊断任务，我们将诊断结论
Regarding the TCM diagnosis task, in order to facilitate quantitative evaluation, we transformed the diagnostic inquiries into five-choice multiple-choice questions. Specifically, we augmented the questions with additional disease options from the same disease group. Subsequently, patient information and disease options were fed as inputs, and the predicted option was extracted from the model-generated response using regular expressions.
% 对于这两个任务，我们都使用准确率来评价。
For both of these tasks, we employed accuracy as the evaluation metric. This involved computing accuracy by comparing the extracted options with the accurate answers.

\subsection{Training Details}
We conducted a series of comprehensive experiments, following \cite{wang2023clinicalgpt}.
We opted to utilize the BLOOM-7B model \cite{scao2022bloom} as our foundational architecture, given its multilingual capabilities and public accessibility.
During the pretraining process, we configured the learning rate to 1e-4, employed a batch size of 128, a maximum sequence length of 1024 tokens, spanned across 2 epochs. 
During the supervised fine-tuning process, we established the learning rate to 5e-5, with a batch size of 128, a maximum sequence length of 1024, training across 3 epochs.
We adopted LoRA with the rank $r$ to 8, the scaling factor to 32, and a dropout rate of 0.1.
\section{Results}
\subsection{Evaluation on Two TCM Tasks}

% \begin{table*}[!h]
% \centering
% \caption{Performance of Models Pretrained with Different Corpus}
% \label{tab:1}
% \begin{tabular}{@{}>{\centering\arraybackslash}lccccccc@{}}
% \toprule
%       &  \thead{TCD}  &  \thead{FS}  & \thead{ES}  &  \thead{SMD}  &  \thead{Examination \\average }  &  \thead{Medical Record \\Diagnosis average}   & \thead{average} \\ \midrule
% Raw & 0.252 & 0.242 & 0.235 & 0.234 & 0.247 & 0.235 & 0.241 \\ 
% Random & 0.278(+10.31\%) & \textbf{0.268(+10.74\%)} & \textbf{0.269(+14.47\%)} & 0.213(-8.97\%) & 0.273(+10.53\%) & 0.241(+2.55\%) & 0.257(+6.64\%) \\
% TCM   & \textbf{0.315(+25.00\%)} & 0.265(+9.50\%) & 0.252(7.23\%) & \textbf{0.277(+18.38\%)} & \textbf{0.290(+17.41\%)} & \textbf{0.264(12.34\%)} & \textbf{0.277(+14.94\%)} \\ \bottomrule  
% \end{tabular}
% \end{table*}

\begin{table}[t]
\centering
\caption{Evaluation Results of TCM Examination Task.}
\label{tab:1}
\begin{tabular}{@{}>{\centering\arraybackslash}lccc@{}}
\toprule
      &  TCD  &  FS &  Average   \\ \midrule
Baseline & 0.252 & 0.242 & 0.247 \\ 
Random & 0.278 (+10.3\%) & \textbf{0.268 (+10.7\%)} & 0.273 (+10.5\%) \\
TCMDA   & \textbf{0.315 (+25.0\%)} & 0.265 (+9.5\%) & \textbf{0.290 (+17.4\%)} \\ \bottomrule  
\end{tabular}
\end{table}

\begin{table}[t]
\centering
\caption{Evaluation Results of TCM Diagnosis Task.}
\label{tab:2}
\begin{tabular}{@{}>{\centering\arraybackslash}lccc@{}}
\toprule
      &  ES & SMD & Average \\ \midrule
Baseline &  0.235 & 0.234 & 0.235 \\ 
Random & \textbf{0.269 (+14.4\%)} & 0.213 (-8.9\%) & 0.241 (+2.5\%) \\
TCMDA   & 0.252 (+7.2\%) & \textbf{0.277 (+18.3\%)} & \textbf{0.264 (+12.3\%)} \\ \bottomrule  
\end{tabular}
\end{table}

% 我们将没有语料预训练的、使用随机语料预训练的、使用中医语料预训练的做对比。
% Baseline
% Random
% TCMDA (ours)
To evaluate the impact of various pre-training strategies on the performance of models in TCM tasks, we conducted a comprehensive evaluation using TCM-EXAM and TCM-EHR datasets. The models encompassed three pre-training strategies, followed by fine-tuning: Baseline, models without any additional pre-training. Random, models pre-trained on a random corpus of size equivalent to TCM-Corpus-1B. TCMDA (ours), models pre-trained on TCM-specific corpus, namely TCM-Corpus-1B.
Tables \ref{tab:1} and \ref{tab:2} present the evaluation results for the TCM examination and diagnosis tasks, respectively. The percentages within parentheses indicate the relative performance compared to the baseline model (Baseline).

% 整体情况，说明预训练是有效的
As shown in Tables \ref{tab:1} and \ref{tab:2}, we found that models with further pretraining consistently outperform those without additional pretraining. Specifically, the Random model exhibited performance improvements of $10.5\%$ and $2.5\%$ on the TCM-EXAM and TCM-EHR tasks, respectively. Our domain-specific TCMDA model showcased more enhancements with increases of $17.4\%$ for TCM-EXAM and $12.3\%$ for TCM-EHR, achieving accuracies of $0.290$ and $0.264$, respectively. 
% 小讨论，提前解惑 随机选语料也有提高，可能因为BLOOM主要在英文上训练，预训练中文适配后性能有提升
Notably, Random exhibited a performance boost with pretraining on a random corpus. This enhancement can possibly be attributed to the fact that BLOOM, our pre-training framework, lacks exposure to Chinese language data, thus providing certain benefits in a different context. However, the most remarkable advancements were observed with the TCMDA model, indicating the effectiveness of domain-specific pre-training.
% 随机语料训练可能下降，而我们模型一致提高
It's worth noting that the Random model exhibited an accuracy decline of $8.9\%$ in the SMD task of TCM-EXAM. Conversely, the TCMDA model consistently outperformed across all tasks, demonstrating that domain-specific pre-training not only enhances performance but also improves model robustness across various scenarios.
% 简单讨论 TCM-EXAM和TCM-EHR的性能差异 体现我们两个任务上都表现很好
The greater enhancement in TCM-EXAM ($17.4\%$) compared to TCM-EHR ($12.3\%$) could be attributed to the broader scope of TCM-EXAM, which benefits more from domain pre-training. While TCM-EHR primarily deals with clinical records. Even though extracting specialized content can be challenging, our domain-specific pre-training significantly improved the model's comprehension of these records.

\subsection{Effect of Corpus Size}

\begin{figure}[t]
  \centering
  \includegraphics[width=0.5\textwidth]{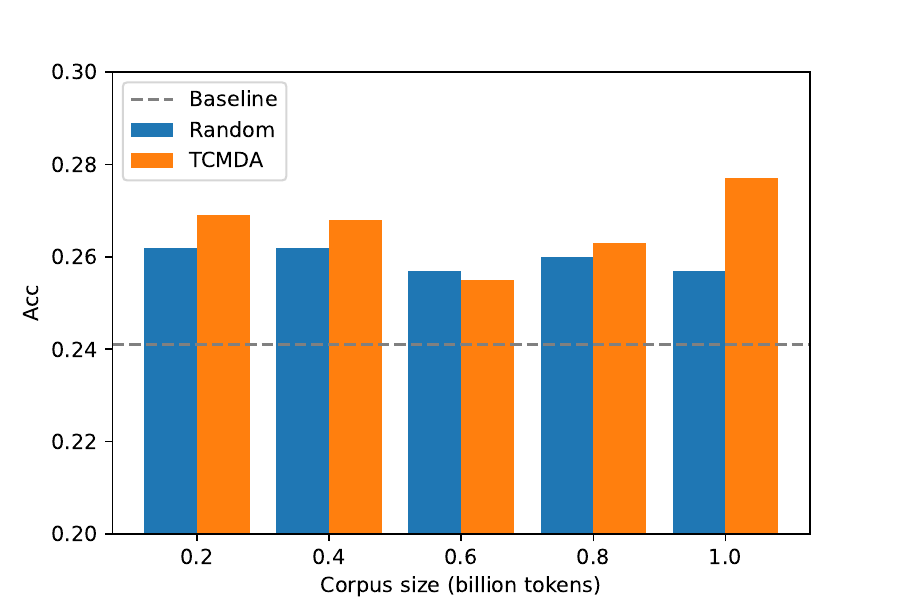} % 替换为你的图片文件名和路径
  \caption{The Impact of Corpus Size on Performance}
  \label{fig:2}
\end{figure}
To study the effect of corpus size on model performance, we conducted experiments with varying amounts of corpus for pretraining, as shown in Figure \ref{fig:2}.
Our TCMDA model, pretrained using TCM-related corpus, consistently demonstrated superior performance compared to Random.
% 随机的增加趋于平。TCM趋于提升，说明随机的很可能就是中文带来的。
% 这说明了随机语料带来了中文领域的一些信息帮助性能固定提升，而中医语料带来的中医领域的一些信息带来的性能提升是有随着预训练语料的增多而变大的趋势的。
As demonstrated in \ref{tab:1} and \ref{tab:2}, the Random exhibited improved Chinese language capabilities over Baseline, as also evidenced in Figure \ref{fig:2}. However, with the expansion of the corpus size, Random did not display a trend of increasing in performance. In contrast, the TCMDA model, pretrained using TCM-related corpus, exhibited a upward trend in performance as the size of the corpus grew. This observation also suggests the rationale for further experimental validation across a broader spectrum of corpora in the future study.

% To study the effect of corpus size on model performance, we also use different amounts of corpus for pretraining and record the results in Figure \ref{fig:2}.
% As shown in Figure \ref{fig:2}, the models pretrained with TCM related corpus almost always perform better than those without pretrained or those pretrained with random corpus.
% % 随机的增加趋于平。TCM趋于提升，说明随机的很可能就是中文带来的。
% As the size of the corpus increased, there is no corresponding improvement in the performance of models pre-trained on random corpus.
% However, models pre-trained on TCM related corpus show a trend of performance improvement.
% % 这说明了随机语料带来了中文领域的一些信息帮助性能固定提升，而中医语料带来的中医领域的一些信息带来的性能提升是有随着预训练语料的增多而变大的趋势的。
% This indicates that random corpus provide some helpful information for fiexed performance improvement in the Chinese domain, while the performance enhancement brought by TCM corpus shows a trend of increase with the growth of pre-training data.
% % The possible reason for this may be related to the total size of our corpus and the model capacity.
% In future work, we will explore larger models and incorporate additional corpora.
\subsection{Case Study}

\begin{table}[t]
\scriptsize
\centering
\begin{CJK*}{UTF8}{gbsn}
\caption{A sample question extracted from TCM-EXAM test set.}
\label{tab:3}
\begin{tabular}{@{}p{8cm}@{}}
\toprule
\textbf{Question}  \\ \midrule
脉在筋骨，乍疏乍密，散乱无序者，称为\\ 
回答选项：A. 鱼翔脉; B. 虾游脉; C. 雀啄脉; \textcolor{red}{D. 解索脉}\\
请分析并给出正确选项。 \\ 
Translation: \\
Pulse in the tendons and bones, sometimes sparse and sometimes dense, scattered and disordered, is called \\
Answer options: A. Waving fish pulse; B. Darting shrimp pulse; C. Pecking sparrow pulse; \textcolor{red}{D. Untwining rope pulse})\\
Please analyze and provide the correct option.
\\ \midrule
\textbf{Response 1 (TCMDA)}
\\ \midrule
脉在筋骨，乍疏乍密，散乱无序者，称为解索脉。解索脉是指脉搏在筋骨之间出现，乍疏乍密，散乱无序的状态。脉搏在筋骨之间出现，意味着脉搏在筋骨之间的血管中流动。乍疏乍密，散乱无序的状态，意味着脉搏在筋骨之间出现时，脉搏的强弱、节律和节奏都不稳定。因此，解索脉是脉搏在筋骨之间出现的一种状态。\textcolor{red}{因此，正确选项是D. 解索脉。}\\
Translation: \\
The pulse in the muscles and bones, sparse and dense at first, scattered and disorderly, is called untwining rope pulse. Untwining rope pulse refers to the state in which the pulse appears between muscles and bones, initially sparse and dense, scattered and disorderly. The pulse appears between the muscles and bones, indicating that it flows through the blood vessels between the muscles and bones. The state of being scattered and disorderly means that the strength, rhythm, and rhythm of the pulse are unstable when it appears between muscles and bones. Therefore, untwining rope pulse is a state in which the pulse appears between muscles and bones. \textcolor{red}{Therefore, the correct option is D Untwining rope pulse.}
\\ \midrule
\textbf{Response 2 (Random)}
\\ \midrule
脉是人体经络系统的一部分，在筋骨中运行。脉的形态和结构非常复杂，乍疏乍密，散乱无序。这种形态和结构的特点与鱼翔脉的形态和结构相似。因此，脉在筋骨，乍疏乍密，散乱无序者，称为鱼翔脉。\textcolor{red}{因此，正确选项是A. 鱼翔脉。}\\
Translation: \\
Pulse is a part of the human meridian system that operates in the muscles and bones. The morphology and structure of veins are very complex, with sparse and dense patterns, scattered and disorderly. The characteristics of this form and structure are similar to those of fish-swimming pulse. Therefore, the pulse in the muscles and bones, which are sparse and dense at first, scattered and disorderly, are called the waving fish pulse. \textcolor{red}{Therefore, the correct option is A Waving fish pulse.}\\
\bottomrule
\end{tabular} 
\end{CJK*}
\end{table}
Table \ref{tab:3} illustrates a task sample extracted from the TCM-EXAM test dataset. 
Response 1 is generated by our TCMDA model, while Response 2 is the output of the Random model. Notably, Response 1 correctly addresses the query, whereas Response 2 falls short in providing the accurate answer. Despite both responses exhibit linguistic fluency, Response 2 lacks meaningful and informative content.

% 有领域知识预训练，模型生成有用的分析，促进得到正确回答
For example, within Response 1, the sentence "The pulse in the muscles and bones, sparse and dense at first, scattered and disorderly, is called untwining rope pulse." effectively describes the definition of untwining rope pulse, with subsequent analysis stemming from this definition leading to the accurate answer. This demonstrates how information gained from the TCM corpus enriches the quality of model-generated responses, thereby augmenting the overall performance of the model.
% 这个结果显示，没有领域知识预训练，模型生成不了有用的信息，并错误回答
In contrast, Response 2's sentence "Pulse is a part of the human meridian system that operates in the muscles and bones. The morphology and structure of veins are very complex, with sparse and dense patterns, scattered and disorderly." merely echoes the original question and subsequently arrives at an incorrect answer. This result shows that without domain-specific knowledge learned through pretraining, models struggle to generate useful insights and provide accurate responses.

\section{Conclusion}
% introduction的最后一段
% To address these challenges, this study introduces an approach aimed at efficiently pre-training models that can be efficiently tailored to the domain of TCM. Our contributions are summarized as follows:
% Firstly, to overcome the limited availability of pre-training data in TCM, we adopted a strategic approach by extracting domain text from general corpus. By incorporating domain terminologies and extracting task-relevant keywords, we constructed a set of keywords specific to TCM. This set was then utilized to retrieve TCM-relevant text, resulting in the TCM-Corpus-1B, a substantial collection of TCM-related textual data.
% Secondly, to ensure the alignment of large language models to the TCM domain, we employed the LoRA technique for domain adaptation during pre-training and fine-tuning, leading to TCM-GPT-7B. The performance of TCM-GPT-7B was evaluated across two distinct TCM tasks: TCM examinations and TCM diagnosis. The experimental results demonstrate our models archive state-of-the-art performance after pre-training using the TCM-Corpus-1B, revealing the necessity and effectiveness of domain-specific pre-training.
% Thirdly, we are committed to publicly releasing both the curated the TCM-Corpus-1B and the TCM-GPT-7B model, as this is intended to boost cross-disciplinary advancements in both TCM and NLP.

In this paper, we have proposed a strategic domain adaptation method for TCM, TCMDA, to overcome the limited availability of pre-training data in TCM. To the best of our knowledge, TCMDA represents the first attempt of domain adaptation of a large language model with 7 billion parameters in TCM domain.
The core idea is to construct a TCM-specic corpus, TCM-Corpus-1B, by retrieving from general corpus and adapt our models on it.
To align our models to the TCM domain, we use the LoRA technique for domain adaptation during pre-training and fine-tuning. Furthermore, we conduct a multi-level evaluation for our approach, including simple TCM examination task and complex real-world TCM diagnosis task. Our experiments show that TCMDA helps to enhance the model performance on various TCM downstream tasks, revealing the necessity and effectiveness of domain-specific pre-training. We will release both TCM-Corpus-1B and TCM-GPT-7B model to facilitate interdisciplinary development in TCM and NLP.

\section*{Acknowledgment}
This study was funded by the National Natural Science Foundation of China (grant 62272055), New Cornerstone Science Foundation through the XPLORER PRIZE, and Young Elite Scientists Sponsorship Program by CAST (2021QNRC001). The authors appreciate Nanjing Institute of InforSuperBahn for providing the traning and evaluation platform.

% \section*{References}
\bibliographystyle{./bib/IEEEtran}
\bibliography{./bib/refs} % 正文无引用时会产生error

\begin{thebibliography}{10}
\providecommand{\url}[1]{#1}
\csname url@samestyle\endcsname
\providecommand{\newblock}{\relax}
\providecommand{\bibinfo}[2]{#2}
\providecommand{\BIBentrySTDinterwordspacing}{\spaceskip=0pt\relax}
\providecommand{\BIBentryALTinterwordstretchfactor}{4}
\providecommand{\BIBentryALTinterwordspacing}{\spaceskip=\fontdimen2\font plus
\BIBentryALTinterwordstretchfactor\fontdimen3\font minus
  \fontdimen4\font\relax}
\providecommand{\BIBforeignlanguage}[2]{{%
\expandafter\ifx\csname l@#1\endcsname\relax
\typeout{** WARNING: IEEEtran.bst: No hyphenation pattern has been}%
\typeout{** loaded for the language `#1'. Using the pattern for}%
\typeout{** the default language instead.}%
\else
\language=\csname l@#1\endcsname
\fi
#2}}
\providecommand{\BIBdecl}{\relax}
\BIBdecl

\bibitem{han2021pre}
X.~Han, Z.~Zhang, N.~Ding, Y.~Gu, X.~Liu, Y.~Huo, J.~Qiu, Y.~Yao, A.~Zhang,
  L.~Zhang \emph{et~al.}, ``Pre-trained models: Past, present and future,''
  \emph{AI Open}, vol.~2, pp. 225--250, 2021.

\bibitem{alaparthi2020bidirectional}
S.~Alaparthi and M.~Mishra, ``Bidirectional encoder representations from
  transformers (bert): A sentiment analysis odyssey,'' \emph{arXiv preprint
  arXiv:2007.01127}, 2020.

\bibitem{radford2018improving}
A.~Radford, K.~Narasimhan, T.~Salimans, I.~Sutskever \emph{et~al.}, ``Improving
  language understanding by generative pre-training,'' 2018.

\bibitem{zhao2023survey}
W.~X. Zhao, K.~Zhou, J.~Li, T.~Tang, X.~Wang, Y.~Hou, Y.~Min, B.~Zhang,
  J.~Zhang, Z.~Dong \emph{et~al.}, ``A survey of large language models,''
  \emph{arXiv preprint arXiv:2303.18223}, 2023.

\bibitem{brown2020language}
T.~Brown, B.~Mann, N.~Ryder, M.~Subbiah, J.~D. Kaplan, P.~Dhariwal,
  A.~Neelakantan, P.~Shyam, G.~Sastry, A.~Askell \emph{et~al.}, ``Language
  models are few-shot learners,'' \emph{Advances in neural information
  processing systems}, vol.~33, pp. 1877--1901, 2020.

\bibitem{chowdhery2022palm}
A.~Chowdhery, S.~Narang, J.~Devlin, M.~Bosma, G.~Mishra, A.~Roberts, P.~Barham,
  H.~W. Chung, C.~Sutton, S.~Gehrmann \emph{et~al.}, ``Palm: Scaling language
  modeling with pathways,'' \emph{arXiv preprint arXiv:2204.02311}, 2022.

\bibitem{gururangan2020don}
S.~Gururangan, A.~Marasovi{\'c}, S.~Swayamdipta, K.~Lo, I.~Beltagy, D.~Downey,
  and N.~A. Smith, ``Don't stop pretraining: Adapt language models to domains
  and tasks,'' \emph{arXiv preprint arXiv:2004.10964}, 2020.

\bibitem{arumae2020empirical}
K.~Arumae, Q.~Sun, and P.~Bhatia, ``An empirical investigation towards
  efficient multi-domain language model pre-training,'' \emph{arXiv preprint
  arXiv:2010.00784}, 2020.

\bibitem{jin2021lifelong}
X.~Jin, D.~Zhang, H.~Zhu, W.~Xiao, S.-W. Li, X.~Wei, A.~Arnold, and X.~Ren,
  ``Lifelong pretraining: Continually adapting language models to emerging
  corpora,'' \emph{arXiv preprint arXiv:2110.08534}, 2021.

\bibitem{shah2022flue}
R.~S. Shah, K.~Chawla, D.~Eidnani, A.~Shah, W.~Du, S.~Chava, N.~Raman,
  C.~Smiley, J.~Chen, and D.~Yang, ``When flue meets flang: Benchmarks and
  large pre-trained language model for financial domain,'' \emph{arXiv preprint
  arXiv:2211.00083}, 2022.

\bibitem{lan2019albert}
Z.~Lan, M.~Chen, S.~Goodman, K.~Gimpel, P.~Sharma, and R.~Soricut, ``Albert: A
  lite bert for self-supervised learning of language representations,''
  \emph{arXiv preprint arXiv:1909.11942}, 2019.

\bibitem{sanh2019distilbert}
V.~Sanh, L.~Debut, J.~Chaumond, and T.~Wolf, ``Distilbert, a distilled version
  of bert: smaller, faster, cheaper and lighter,'' \emph{arXiv preprint
  arXiv:1910.01108}, 2019.

\bibitem{hu2021lora}
E.~J. Hu, Y.~Shen, P.~Wallis, Z.~Allen-Zhu, Y.~Li, S.~Wang, L.~Wang, and
  W.~Chen, ``Lora: Low-rank adaptation of large language models,'' \emph{arXiv
  preprint arXiv:2106.09685}, 2021.

\bibitem{lee2020biobert}
J.~Lee, W.~Yoon, S.~Kim, D.~Kim, S.~Kim, C.~H. So, and J.~Kang, ``Biobert: a
  pre-trained biomedical language representation model for biomedical text
  mining,'' \emph{Bioinformatics}, vol.~36, no.~4, pp. 1234--1240, 2020.

\bibitem{BioMedLM2023}
\BIBentryALTinterwordspacing
M.~C. Abhinav~Venigalla, Jonathan~Frankle. (2023) Biomedlm: a domain-specific
  large language model for biomedical text. [Online]. Available:
  \url{https://www.mosaicml.com/blog/introducing-pubmed-gpt}
\BIBentrySTDinterwordspacing

\bibitem{tang2008traditional}
J.-L. Tang, B.-Y. Liu, and K.-W. Ma, ``Traditional chinese medicine,''
  \emph{The Lancet}, vol. 372, no. 9654, pp. 1938--1940, 2008.

\bibitem{zhang2020constructing}
T.~Zhang, Y.~Wang, X.~Wang, Y.~Yang, and Y.~Ye, ``Constructing fine-grained
  entity recognition corpora based on clinical records of traditional chinese
  medicine,'' \emph{BMC medical informatics and decision making}, vol.~20,
  no.~1, pp. 1--17, 2020.

\bibitem{yin2023large}
Z.~Yin, Q.~Sun, Q.~Guo, J.~Wu, X.~Qiu, and X.~Huang, ``Do large language models
  know what they don't know?'' \emph{arXiv preprint arXiv:2305.18153}, 2023.

\bibitem{he2022rethinking}
H.~He, H.~Zhang, and D.~Roth, ``Rethinking with retrieval: Faithful large
  language model inference,'' \emph{arXiv preprint arXiv:2301.00303}, 2022.

\bibitem{mihalcea2004textrank}
R.~Mihalcea and P.~Tarau, ``Textrank: Bringing order into text,'' in
  \emph{Proceedings of the 2004 conference on empirical methods in natural
  language processing}, 2004, pp. 404--411.

\bibitem{robertson2009probabilistic}
S.~Robertson, H.~Zaragoza \emph{et~al.}, ``The probabilistic relevance
  framework: Bm25 and beyond,'' \emph{Foundations and Trends{\textregistered}
  in Information Retrieval}, vol.~3, no.~4, pp. 333--389, 2009.

\bibitem{yao2022nlp}
X.~Yao, Y.~Zheng, X.~Yang, and Z.~Yang, ``Nlp from scratch without large-scale
  pretraining: A simple and efficient framework,'' in \emph{International
  Conference on Machine Learning}.\hskip 1em plus 0.5em minus 0.4em\relax PMLR,
  2022, pp. 25\,438--25\,451.

\bibitem{wang2023clinicalgpt}
G.~Wang, G.~Yang, Z.~Du, L.~Fan, and X.~Li, ``Clinicalgpt: Large language
  models finetuned with diverse medical data and comprehensive evaluation,''
  \emph{arXiv preprint arXiv:2306.09968}, 2023.

\bibitem{scao2022bloom}
T.~L. Scao, A.~Fan, C.~Akiki, E.~Pavlick, S.~Ili{\'c}, D.~Hesslow,
  R.~Castagn{\'e}, A.~S. Luccioni, F.~Yvon, M.~Gall{\'e} \emph{et~al.},
  ``Bloom: A 176b-parameter open-access multilingual language model,''
  \emph{arXiv preprint arXiv:2211.05100}, 2022.

\end{thebibliography}

\end{document}